\pdfoutput=1
\documentclass[a4paper,fleqn]{cas-dc}

\usepackage[numbers]{natbib}
\usepackage{enumitem}
\usepackage[frozencache=true,cachedir=.]{minted}
\usepackage{pdfpages}

\begin{document}
\let\WriteBookmarks\relax
\def\floatpagepagefraction{1}
\def\textpagefraction{.001}

\shorttitle{MLHarness: A Scalable Benchmarking System for MLCommons}    

\shortauthors{Y. Chang et al.}

\title [mode = title]{MLHarness: A Scalable Benchmarking System for MLCommons}  



\affiliation[1]{organization = University of Illinois at Urbana-Champaign, city = Urbana, state = IL, country = USA}
\affiliation[2]{organization = University at Buffalo, city = Buffalo, state = NY, country = USA}
\author[1]{Yen-Hsiang Chang}
\ead[]{yhchang3@illinois.edu}
\author[1]{Jianhao Pu}
\ead[]{jpu3@illinois.edu}
\author[1]{Wen-mei Hwu}
\ead[]{w-hwu@illinois.edu}
\author[1, 2]{Jinjun Xiong}
\ead[]{jinjunx@illinois.edu}
\ead[]{ jinjun@buffalo.edu}
















\begin{abstract}
With the society's growing adoption of machine learning (ML) and deep learning (DL) for various intelligent solutions, 
it becomes increasingly imperative to standardize a common set of measures for ML/DL models with large scale open datasets under common development practices and resources so that people can benchmark and compare models quality and performance on a common ground. MLCommons has emerged recently as a driving force from both industry and academia to orchestrate such an effort. Despite its wide adoption as standardized benchmarks, MLCommons Inference has only included a limited number of ML/DL models (in fact seven models in total). This significantly limits the generality of MLCommons Inference's benchmarking results because there are many more novel ML/DL models from the research community, solving a wide range of problems with different inputs and outputs modalities. To address such a limitation, we propose MLHarness, a scalable benchmarking harness system for MLCommons Inference with three distinctive features: 
(1) it codifies the standard benchmark process as defined by MLCommons Inference 
including the models, datasets, DL frameworks, and software and hardware systems; 
(2) it provides an easy and declarative approach for model developers to contribute their models and datasets to MLCommons Inference; and
(3) it includes the support of a wide range of models with varying inputs/outputs modalities so that we can scalably benchmark these models across different datasets, frameworks, and hardware systems. This harness system is developed on
top of the MLModelScope system, and will be
open sourced to the community. Our experimental results demonstrate the superior flexibility and scalability of this harness system for MLCommons Inference benchmarking.
\end{abstract}




\maketitle










\newcommand{\SYSNAME}{MLHarness}

\section{Introduction}

With the rise of machine learning (ML) and deep learning (DL) innovations in both industry and academia, there is a clear need for standardized benchmarks and evaluation criteria to facilitate comparison and development of ML/DL innovations. MLCommons Inference \cite{reddi2019mlperf}, a standard ML/DL inference benchmark suite with properly defined metrics and benchmarking methodologies, has emerged recently to facilitate such an effort. However, MLCommons Inference only included five models when it was first introduced in 2019, and has only included seven models recently \cite{MLCommonsInferenceGithub}. This limited and slow-growing number of ML/DL models stifles the adoption of MLCommons Inference as a general benchmarking platform because there are many more novel ML/DL models from the research community, solving a wide range of problems with different inputs and outputs modalities. To address such a limitation, we propose \SYSNAME{}, a scalable benchmarking harness system for MLCommons Inference, to make MLCommons Inference embrace new models and modalities easily. \SYSNAME{} is developed on top of MLModelScope \cite{DBLP:journals/corr/abs-2002-08295} with three distinctive new features: (1) it codifies the required benchmarking environment for MLCommons Inference explicitly, including models, datasets, frameworks, software and hardware stacks; (2) it provides an easy and declarative approach for model developers to contribute their models and datasets to MLCommons Inference; and (3) it supports a wide range of models with varying inputs and outputs modalities. Our experiments show that \SYSNAME{} is capable of reporting all the required metrics as defined by MLCommons Inference for models with different input/output modalities and for models both within and beyond MLCommons Inference.

In particular, we make the following contributions: (1) we propose \SYSNAME{}, a scalable benchmarking harness system for MLCommons Inference while supporting models beyond those in MLCommons Inference; (2) we extend
MLModelScope to provide
 user-defined pre-processing and post-processing interfaces so that MLModelScope can easily support new models and new modalities; (3) we showcase \SYSNAME{}' capabilities as a scalable benchmarking harness system for MLCommons Inference by running experiments on a range of models both within and beyond MLCommons Inference under different frameworks and systems configurations; and (4) we further demonstrate the unique value of scalable benchmarking in identifying abnormal system
 behaviors and how  \SYSNAME{} helps to explain those seemingly
 abnormal behaviors resulting from complex software and hardware interactions.

\section{Background}

\subsection{ML/DL Benchmark Challenges}
ML and DL innovations such as applications, datasets, frameworks, models, and software and hardware systems, are being developed in a rapid pace. However, current practice of sharing ML/DL innovations is to build ad-hoc scripts and write manuals to describe the workflow. This makes it hard to reproduce the reported metrics and to port the innovations to different environments and solutions. Therefore, having a standard benchmarking platform with an exchange specification and well-defined metrics to fairly compare and benchmark the innovations is a crucial step toward the success of ML/DL community.


Previous work includes (1) ML/DL model zoos curated by framework developers \cite{GluonCV,ModelHub,ModelZoo,ONNXModelZoo,TFHub,TorchVision}, but they only aim for sharing ML/DL models as a library; (2) package managers for a specific software environment such as Spack \cite{Spack}, while they are just targeting on maintaining packages in different software and hardware stacks; (3) benchmarking platforms such as MLCommons \cite{mattson2019mlperf,reddi2019mlperf} and MLModelScope \cite{DBLP:journals/corr/abs-2002-08295}, but the former only focuses on few specific models and the latter only focuses on models in computer vision tasks; (4) collections of reproducible MLOps components and architectures \cite{CloudOps,DBLP:journals/corr/abs-2011-01149,MLOps}, while their main focuses are on deployment and automation; (5) plug-and-play shareable containers such as MLCube \cite{MLCube}, but its generality makes it hard to identify and locate the crucial components for the cause of abnormal behaviors in ML/DL models; (6) simulator of ML/DL inference servers such as iBench \cite{iBench}, but the main focus on capturing data transfer capabilities between clients and servers provides no insights on profiling models. As the above applications either only focus on a specific software and hardware stack, or use ad-hoc approaches to handle specific ML/DL tasks, or are lack of a consistent benchmarking method, it is hard to use them individually to have a well rounded experience when developing ML/DL innovations.

To address these ML/DL benchmark challenges, we propose a new scalable benchmarking system: \SYSNAME{} by taking advantage of two open-source projects: MLModelScope \cite{DBLP:journals/corr/abs-2002-08295} for its exchange specification on software and hardware stacks, and MLCommons Inference \cite{reddi2019mlperf} for its community-adopted benchmarking scenarios and metrics. With \SYSNAME{}, we are able to benchmark and compare quality and performance of models on a common ground through a set of well-defined metrics and exchange specification.

\subsection{Overview of MLCommons}
MLCommons \cite{mattson2019mlperf,reddi2019mlperf}, previously known as MLPerf, is a platform aims to answer the needs of the nascent machine learning industry. MLCommons Training \cite{mattson2019mlperf} measures how fast systems can train models to a target quality metric, while MLCommons Inference \cite{reddi2019mlperf} measures how fast systems can process inputs and produce results using a trained model. Both of these two benchmark suites target on providing benchmarking results on different scales of computing services, ranging from tiny mobile devices to high performance computing data centers. As the main focus of this paper is on benchmarking ML/DL inferences, we only focus on MLCommons Inference in the rest of this paper.

\subsubsection{Characteristics of MLCommons Inference}
MLCommons Inference is a standard ML/DL inference benchmark suite with a set of properly defined metrics and benchmarking methodologies to fairly measure the inference performance of ML/DL hardware, software, and services. MLCommons Inference focuses on the following perspectives when designing its benchmarking metrics:
\begin{itemize}
    \item \textbf{Selection of Representative Models.} MLCommons Inference selects representative models that are mature, open source, and have earned community support. This permits accessibility and reproducible measurements, which facilitates MLCommons Inference becoming a standardized benchmarking suite.
    \item \textbf{Scenarios.} MLCommons Inference consists of four evaluation scenarios, including single-stream, multistream, server, and offline. These four scenarios aim for simulating realistic behaviors of inference systems in many critical applications.
    \item \textbf{Metrics.} Apart from the commonly used model metrics such as accuracy, MLCommons Inference also includes a set of systems related metrics, such as percentile-latency and throughput. These make MLCommons Inference appealing in satisfying the demand of different use cases, such as 99\% percentile-latency for a data center to respond to a user query.
\end{itemize}
\subsubsection{Workflows of MLCommons Inference}
\begin{figure}[t]
    \begin{center}
        \includegraphics[scale=0.25]{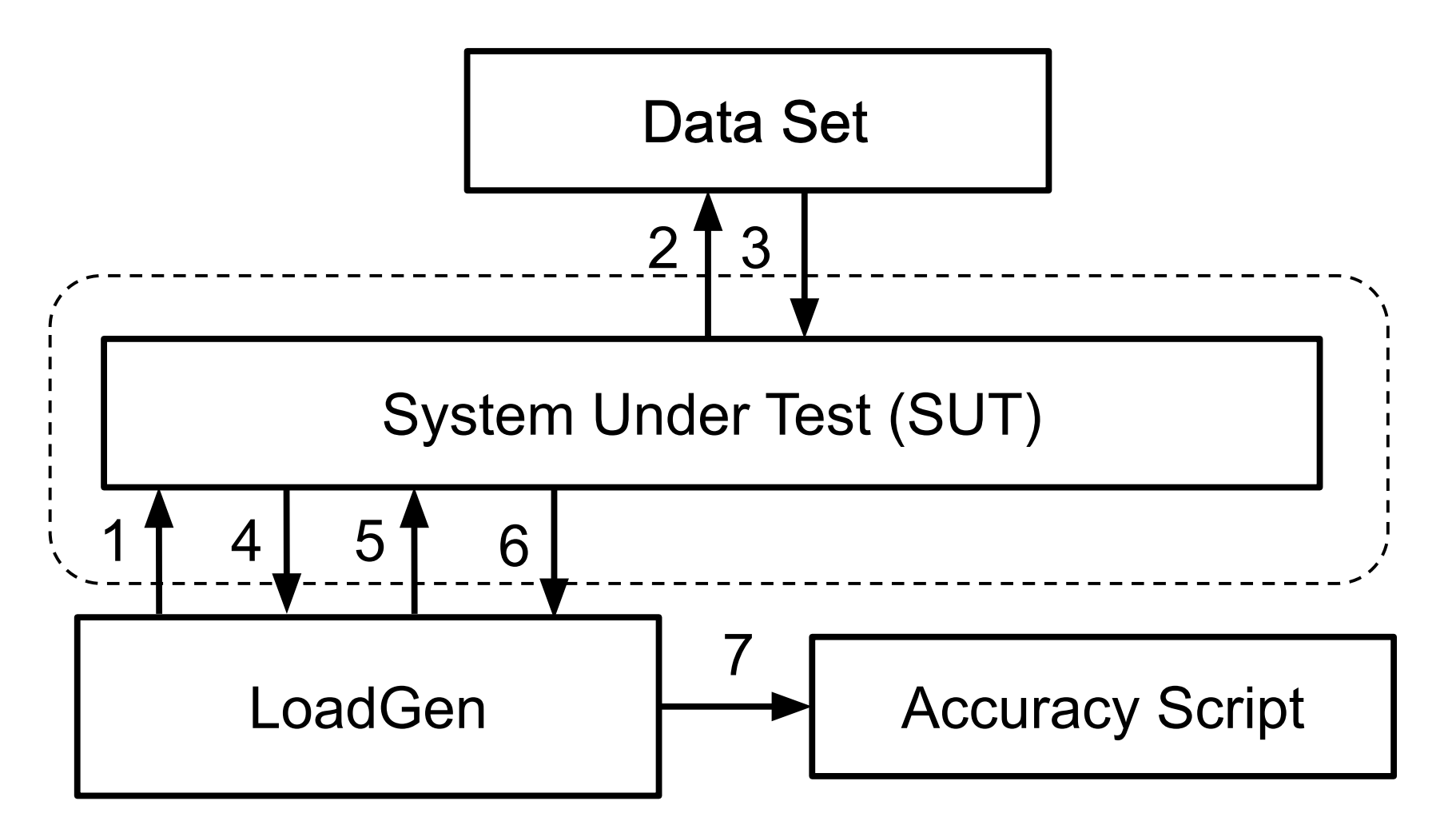}
    \end{center}
    \caption{Workflow of MLCommons Inference \cite{reddi2019mlperf}}
    \label{fig:mlcommonsworkflows}
\end{figure}

Figure \ref{fig:mlcommonsworkflows} shows the critical components as defined in MLCommons Inference, where the numbers and arrows denote the sequence and the directions of the data flows, respectively. The description of the components follows:
\begin{itemize}
    \item \textbf{Load Generator (LoadGen).} The LoadGen produces query traffics as defined by the four scenarios above, collects logging information, and summarizes benchmarking results. It is a stand-alone module that stays the same across different models.
    \item \textbf{System Under Test (SUT).} The SUT consists of the inference system under benchmarking, including ML/DL frameworks, ML/DL models, software libraries and the target hardware system. Once the SUT receives a query from the LoadGen, it completes an inference run and reports the result to the LoadGen.
    \item \textbf{Data Set.} Before issuing queries to the SUT, the LoadGen needs to let the SUT fetch the data needed for the queries from the dataset and pre-process the data. This is not included in the latency measurement.
    \item \textbf{Accuracy Script.} After all queries are issued and the results are received, the accuracy script will be invoked to validate the accuracy of the model from the logging information.
\end{itemize}

\subsubsection{Limitations of MLCommons Inference}
As we can observe from the characteristics and the workflows of MLCommons Inference above, MLCommons Inference involves benchmarking under different scenarios with various metrics, which provides a community acknowledged ML/DL benchmark standard. However, the focus on the seven representative models shadows its advantage because MLCommons Inference only provides ad-hoc scripts for these representative models, and it is hard to extend them to many other models beyond MLCommons Inference.

In fact, the only critical component in MLCommons Inference is the LoadGen, while the other components can be replaced with any inference systems. In this paper, we present how to replace the components other than the LoadGen by MLModelScope \cite{DBLP:journals/corr/abs-2002-08295}, an inference platform with a clearly defined exchange specification and an across-stack profiling and analysis tool, and extend MLModelScope so that it becomes a scalable benchmarking harness for MLCommons Inference. This greatly extends the applicability of MLCommons Inference for models well beyond it. 

\subsection{Overview of MLModelScope}
MLModelScope \cite{DBLP:journals/corr/abs-2002-08295} is a hardware and software agnostic distributed platform for benchmarking and profiling ML/DL models across datasets, frameworks and systems.
\subsubsection{Characteristics of MLModelScope}
\begin{figure}[t]
    \begin{center}
        \includegraphics[scale=0.20]{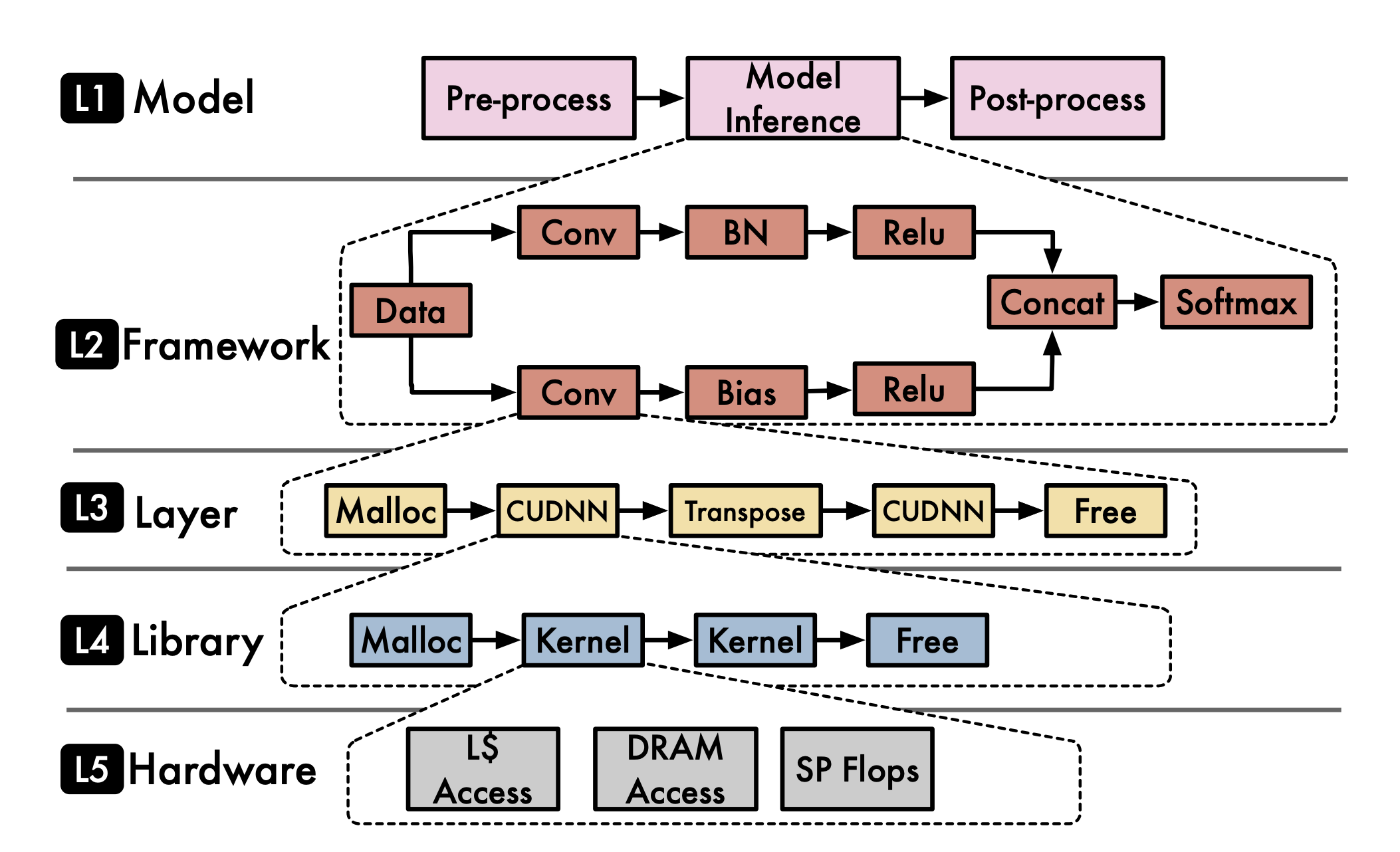}
    \end{center}
    \caption{Profiling levels in MLModelScope \cite{DBLP:journals/corr/abs-2002-08295}}
    \label{fig:level}
\end{figure}

 MLModelScope consists of a specification and a runtime that enable repeatable and fair evaluation. The design aspects follow:
\begin{itemize}
    \item \textbf{Specification.} MLModelScope utilizes the software and model manifests as proposed in DLSpec \cite{DBLP:journals/corr/abs-2002-11262}, which capture different aspects of an ML/DL task and ensure usability and reproducibility. The software manifest defines the software requirements, such as ML/DL frameworks to run an ML/DL task. The model manifest defines the logic to run the model for the ML/DL task, such as pre-processing and post-processing methods, and the required artifact sources. An example is shown in Listing \ref{listing:mlmodelscopemanifest}.
    \item \textbf{Runtime.} The runtime of MLModelScope follows the manifests to set up the required environment for inference. Moreover, MLModelScope includes the across-stack profiling and analysis tool, XSP \cite{DBLP:journals/corr/abs-1908-06869}, which introduces a leveled and iterative measurement approach to overcome the impact of profiling overhead. As shown in Figure \ref{fig:level}, MLModelScope captures profiling data for different levels, which enables users to correlate the information and analyze the performance data in different levels.

\end{itemize}

\begin{listing}[t]
\begin{minted}
[
frame=lines,
framesep=2mm,
baselinestretch=1.0,
fontsize=\footnotesize,
linenos
]
{yaml}
name: Inception-v3 # model name
version: 1.0.0 # semantic version of the model
task: classification
framework: # framework information
    name: TensorFlow 
    version: ^1.x # framework version constraint
model: # model sources
    graph_path: https://.../inception_v3.pb
    graph_checksum: 328f68...3a813e
steps: # pre-processing steps
    decode:
        element_type: int8
        data_layout: NHWC
        color_layout: RGB
    crop:
        method: center
        percentage: 87.5
    resize:
        dimensions: [3, 299, 299]
        method: bilinear
        keep_aspect_ratio: true
    mean: [127.5, 127.5, 127.5]
    rescale: 127.5
...
\end{minted}
\caption{An excerpt of manifest from MLModelScope \cite{DBLP:journals/corr/abs-2002-08295}}
\label{listing:mlmodelscopemanifest}
\end{listing}

\subsubsection{Limitations of MLModelScope}
Although MLModelScope involves a clearly defined specification and is able to run several hundreds of models in different ML/DL frameworks, it currently only supports models for computer vision tasks. While MLModelScope discussed the possibility of using user-defined pre-processing and post-processing inline Python scripts to serve as a universal handler for all kinds of models, MLModelScope did not implement those interfaces but only introduced built-in image manipulations to support computer vision tasks. In this paper, we have actually implemented the user-defined pre-processing and post-processing interfaces and demonstrated its usage on models with different modalities and different pre-processing and post-processing, such as question answering and medical 3D image segmentation.   

\section{\SYSNAME{} Implementation}
This section describes the crucial implementations of \SYSNAME{}, a scalable benchmarking harness system for MLCommons Inference \cite{reddi2019mlperf} for tackling  the limitations of MLCommons Inference and MLModelScope \cite{DBLP:journals/corr/abs-2002-08295}. The implementations include the support of user defined pre-processing and post-processing interfaces and the encapsulation of MLModelScope for MLCommons Inference.

\subsection{Pre-processing and Post-processing Interfaces}
As described in DLSpec \cite{DBLP:journals/corr/abs-2002-11262} and MLModelScope \cite{DBLP:journals/corr/abs-2002-08295}, to make the user defined pre-processing and post-processing interfaces universal, inline Python scripts are chosen to allow great flexibility and productivity, as Python functions can download and run Bash scripts and some C++ code. On the other hand, MLModelScope is implemented in Go; therefore, it is necessary to build a bridge between the runtime of MLModelScope and the embedded Python scripts in the model manifest so that, within the MLModelScope runtime, we can invoke the Python runtime to execute user defined pre-processing and post-processing functions.

A naive solution is to save the input data and the functions as files and execute pre-processing and post-processing functions apart from MLModelScope. However, this approach is impractical since it introduces high serialization and process initialization overhead, and it also makes MLModelScope incapable of supporting streaming data \cite{DBLP:journals/corr/abs-2002-11262}.

In order to avoid using intermediate files, we instead use Python/C APIs \cite{CAPI} to embed a Python interpreter into MLModelScope to execute Python functions, as suggested by DLSpec. To use these APIs, we need to implement wrappers in Go to call them. Instead of building these wrappers from scratch, we use the open source Go-Python3 bindings \cite{DataDog}. In this fashion, the Python functions can be executed within \SYSNAME{} directly to avoid the problems mentioned in the naive solution.

\subsubsection{Implementation Details}

\begin{figure}[t]
    \begin{center}
        \includegraphics[scale=0.083]{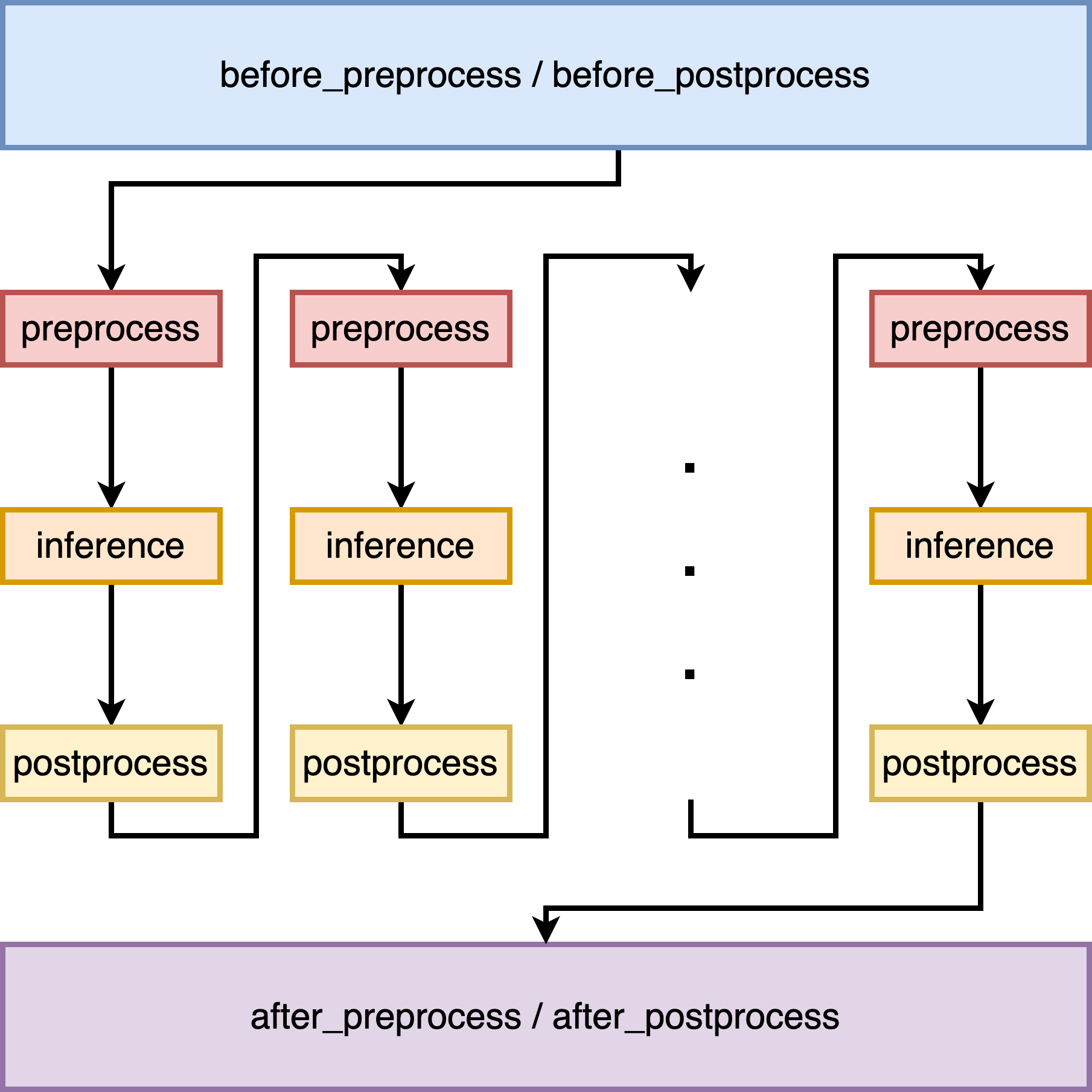}
    \end{center}
    \caption{Workflow of \SYSNAME{}}
    \label{fig:interfaces}
\end{figure}

Figure \ref{fig:interfaces} shows the invocation sequence of user defined pre-processing and post-processing functions by \SYSNAME{}. The \texttt{before\_preprocess} and the \texttt{before\_postprocess} functions are invoked only once at the startup stage; the \texttt{after\_preprocess} function and the \texttt{after\_postprocess} functions are invoked only once after all inferences are done. These four functions are for the sake of loading datasets, writing logging information to files, and specifying configurations during runtime if necessary. The \texttt{preprocess} function and the \texttt{postprocess} function are invoked right before and after every model inference, respectively, to pre-process and post-process the inputs and outputs of the model.

\begin{listing}[t]
\begin{minted}
[
frame=lines,
framesep=2mm,
baselinestretch=1.0,
fontsize=\footnotesize,
linenos
]
{go}
func Processing(tensor interface{}, functionName string) interface{} {
    pyData := MoveDataToPythonInterpreter(tensor)
    pyFunc := FindTheProcessingFunctionByItsName(functionName)
    pyResult := ExecuteProcessingFunction(pyFunc, pyData)
    result := GetResultFromPythonInterpreter(pyResult)
    return result
}
\end{minted}
\caption{Pre-processing and post-processing implementation in Go}
\label{listing:prepostimplementation}
\end{listing}

To embed a Python interpreter, we need to initialize it through Python/C APIs at the beginning of \SYSNAME{}. Then, as Listing \ref{listing:prepostimplementation} shows, the function handling the embedded Python pre-processing and post-processing scripts consists of four parts, utilizing the Go-Python3 bindings:
\begin{itemize}
    \item \texttt{MoveDataToPythonInterpreter.} Moving data from Go to Python is not easy since the data being processed are large, for example, a tensor representing an image. One solution is to serialize the data at one end, transfer the data as a string, and deserialize at the other end. However, it introduces a high overhead due to the high cost of encoding and decoding. To overcome this problem and to make data transfer efficient, we propose to copy the data in-memory, i.e., we only send the shape of the tensor and the address of its underlying flattened array, and reconstruct the tensor by copying data from the address and by its shape. Note that to guarantee the validity of data transfer, we need to make sure that the underlying flattened array represents the tensor contiguously, particularly in case lazy operations were done on the tensor, such as transposition.
    \item \texttt{FindTheProcessingFunctionByItsName.} The processing functions in the model manifest are registered in the \texttt{\_\_main\_\_} module of the Python interpreter during its initialization. To get the corresponding \texttt{PyObject} of these functions, we query the \texttt{\_\_main\_\_} module by the names of functions, which are the six processing functions as listed in Figure \ref{fig:interfaces}.
    \item \texttt{ExecuteProcessingFunction.} The signatures of the processing functions in the model manifest are in the form of \texttt{process(ctx, data)}, where \texttt{ctx} is a dictionary capturing additional information in the manifest, and \texttt{data} is the tensor we got from \texttt{MoveDataToPythonInterpreter}. Therefore, in order to invoke the processing function, we need to call the Go-Python3 binding with the \texttt{PyObject} of the processing function, the dictionary of \texttt{ctx} and the \texttt{data} going to be processed. Note that the \texttt{data} is only effective for \texttt{preprocess} and \texttt{postprocess}, and it is just a \texttt{PyObject} of \texttt{None} for the other four processing functions. 
    \item \texttt{GetResultFromPythonInterpreter.} This is similar to the first part except that it moves data from Python to Go instead of the other way around. Note that we still copy the data in-memory to avoid unnecessary overhead.
\end{itemize}

\begin{figure}[t]
    \begin{center}
        \includegraphics[scale=0.083]{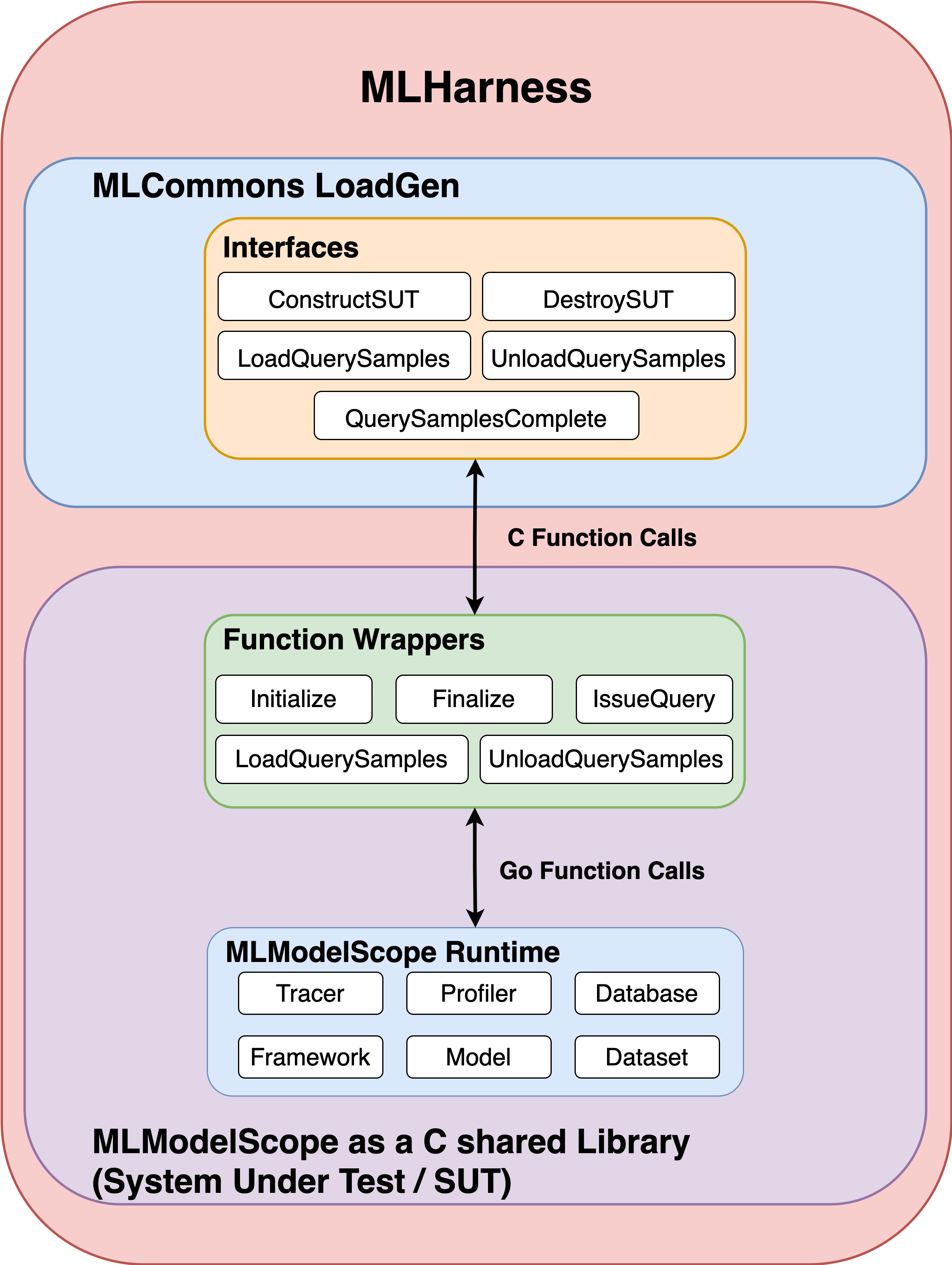}
    \end{center}
    \caption{Structure of \SYSNAME{}}
    \label{fig:struct}
\end{figure}

\subsection{Structure of \SYSNAME{}}
Figure \ref{fig:struct} shows the encapsulation of MLModelScope for MLCommons Inference. In order to utilize MLCommons Inference defined scenarios and performance metrics, we keep the LoadGen and the accuracy script the same as they were in MLCommons Inference. On the other hand, We replace the built-in SUT and data set in MLCommons Inference with MLModelScope runtime to run the models. In this way, MLModelScope is capable of acting as an easy-to-use black box to respond to the queries issued by the LoadGen in MLCommons Inference, and it also provides across-stack profiling information for further analysis, which are not available when merely using MLCommons Inference.

\subsubsection{Implementation Details}
MLModelScope is developed in Go, but the LoadGen in MLCommons Inference is developed in C++ and used in Python through Python bindings. In order to make the communication between MLModelScope and MLCommons Inference feasible, we build MLModelScope as a C shared library \cite{SharedLibrary}, use the ctypes module \cite{Ctypes} in Python to load the shared library, and call the functions in the shared library. Three notable implementations are described below:

\begin{listing}[t]
\begin{minted}
[
frame=lines,
framesep=2mm,
baselinestretch=1.0,
fontsize=\footnotesize,
linenos
]
{yaml}
name: MLPerf_BERT # model name
version: 1.0.0 # semantic version of the model
framework: # framework information
    name: PyTorch 
    version: '>=1.5.0' # framework version constraint
inputs: # model inputs
    - type: text # input modality
      element_type: string
outputs: # model outputs
    - type: text # output modality
      element_type: string
model: # model sources
    graph_path: https://.../bert.pt
    graph_checksum: c3bb5a...aa1ccd
preprocess: |
  from transformers import BertTokenizer
  import numpy as np
  ...
  class SquadExample(object):
    ...
  class InputFeatures(object):
    ...
  def read_squad_examples(...):
    ...
  def convert_examples_to_features(...):
    ...
  features = []
  tokenizer = BertTokenizer(...)
  examples = read_squad_examples(...)
  convert_examples_to_features(features, examples, tokenizer, ...)
  def preprocess(ctx, data):
    cur = features[int(data)]
    return cur.input_ids, cur.input_mask, cur.segment_ids
postprocess: |
  import numpy as np
  import json
  def postprocess(ctx, data):
    res = np.stack([data[0], data[1]], axis = -1).squeeze(0).tolist()
    return [json.dumps(res)]
...
\end{minted}
\caption{An excerpt of model manifest for BERT}
\label{listing:bertmanifest}
\end{listing}

\begin{itemize}
    \item \textbf{Function wrappers.} To simplify the process of building the C shared library and leaving MLModelScope as a black box, we create function wrappers for critical applications in MLModelScope and only export them in the shared library. This includes the \texttt{Initialize} and \texttt{Finalize} wrappers to initialize and finalize the profiling tools in MLModelScope. It also includes the \texttt{LoadQuerySamples}, \texttt{IssueQuery}, and \texttt{UnloadQuerySamples} wrappers to pre-load and pre-process the data from the data set, handle queries from the LoadGen, and free the memory occupied by the pre-loaded data, respectively.
    \item \textbf{Data transmissions.} It is hard to directly exchange data between Go and Python, since there is no one-to-one correspondence between data types in these two languages. To solve this problem, we utilize the built-in primitive C compatible data types in ctypes \cite{Ctypes} for Python and CGO \cite{CGO} for Go, since they define how to transform data if there is no clear correspondence between data types in C and the corresponding languages. Using this method, the data conversion can be done in-memory instead of through serialization.
    \item \textbf{Blocking statements.} When we exchange data between Go and Python, the garbage collector at one end doesn't automatically know that it needs to keep the data before the data are really copied or used at the other end, which might result into undefined behaviors. To solve this problem, we need to manually create blocking statements to block garbage collection until a deep copy of the data is made at the other end. This can be done using the \texttt{KeepAlive} function \cite{KeepAlive} in Go and managing reference counts \cite{RefCnt} in Python to prevent garbage collection being invoked until the \texttt{KeepAlive} is executed and the reference count is decreased to zero, respectively.
\end{itemize}

\subsection{Example of \SYSNAME{}}
With the help of user defined pre-processing and post-processing interfaces, \SYSNAME{} is able to handle various models' inputs and outputs modalities that are not supported in MLModelScope. Also, it is easy to use the model manifest to add models for MLModelScope to report MLCommons Inference defined metrics, which is hard when merely using MLCommons Inference. Listing \ref{listing:bertmanifest} is the model manifest using the pre-processing and post-processing interfaces for BERT \cite{DBLP:journals/corr/abs-1810-04805}, a language representation model, to handle the question answering modality that was not supported in MLModelScope. The pre-processing step uses the tokenizer from the transformers Python third-party library \cite{wolf-etal-2020-transformers} to parse data and prepare input features. The post-processing step reshapes the outputs into the format as defined by the accuracy script. The tedious implementation of the tokenizer is one of the reason why MLModelScope can not support the question answering modality, since it is hard to create an equivalent built-in alternative inside MLModelScope using Go. On the contrary, through the user defined pre-processing and post-processing interfaces, \SYSNAME{} can utilize the community developed Python third-party libraries to overcome this obstacle.

\section{Experimental Results}

\newcommand{\SYSA}{9800-ORT-RTX}
\newcommand{\SYSB}{9800-PT-RTX}
\newcommand{\SYSC}{9800-ORT}
\newcommand{\SYSD}{9800-PT}
\newcommand{\SYSE}{7820-ORT-TITAN}
\newcommand{\SYSF}{7820-PT-TITAN}
\newcommand{\SYSG}{7820-TF}
\newcommand{\SYSH}{AMD-ORT-A100}
\newcommand{\SYSI}{AMD-ORT-V100}
\newcommand{\SYSJ}{9800-MX-RTX}

We conduct two sets of experiments to demonstrate the success of \SYSNAME{} on overcoming the limitations in MLModelScope \cite{DBLP:journals/corr/abs-2002-08295} and MLCommons Inference \cite{reddi2019mlperf}. In the first set of experiments, we use \SYSNAME{} to benchmark models in MLCommons Inference and report MLCommons Inference defined metrics to show that it supports modalities that are not supported in MLModelScope, such as question answering and medical image 3D segmentation. In addition, we show that \SYSNAME{} is able to report the results for all four scenarios defined by MLCommons Inference. In the second set of experiments, we use \SYSNAME{} to benchmark models beyond MLCommons Inference and report MLCommons Inference defined metrics to show the usage of our newly extended MLModelScope as an easy-to-use black box for MLCommons Inference.

\subsection{Experiment Setup}

\begin{table*}[h]
    \centering
    \caption{Systems used for experiments}
    \scriptsize
    \begin{tabular}{|c|c|c|c|}
    \hline
    System  Annotations& Framework & Processor & Accelerator \\
    \hline
    \SYSA{} & ONNX Runtime & 1x Intel(R) Core(TM) i7-9800X CPU @ 3.80GHz & 1x GeForce RTX 3090 \\
    \hline
    \SYSB{} & PyTorch & 1x Intel(R) Core(TM) i7-9800X CPU @ 3.80GHz & 1x GeForce RTX 3090 \\
    \hline
    \SYSC{} & ONNX Runtime & 1x Intel(R) Core(TM) i7-9800X CPU @ 3.80GHz & None \\
    \hline
    \SYSD{} & PyTorch & 1x Intel(R) Core(TM) i7-9800X CPU @ 3.80GHz & None \\
    \hline
    \SYSE{} & ONNX Runtime & 1x Intel(R) Core(TM) i7-7820X CPU @ 3.60GHz & 1x TITAN V \\
    \hline
    \SYSF{} & PyTorch & 1x Intel(R) Core(TM) i7-7820X CPU @ 3.60GHz & 1x TITAN V \\
    \hline
    \SYSG{} & TensorFlow & 1x Intel(R) Core(TM) i7-7820X CPU @ 3.60GHz & None \\
    \hline
    \SYSH{} & ONNX Runtime & 1x AMD EPYC 7702 64-Core Processor & 1x A100 \\
    \hline
    \SYSI{} & ONNX Runtime & 1x AMD EPYC 7702 64-Core Processor & 1x V100 \\
    \hline
    \SYSJ{} & MXNet & 1x Intel(R) Core(TM) i7-9800X CPU @ 3.80GHz & 1x GeForce RTX 3090 \\
    \hline
    \end{tabular}
    \label{tab:sys}
\end{table*}

\begin{table*}[t]
    \centering
    \caption{\SYSNAME{} and MLCommons reported results for all four scenarios on \SYSA{}}
    \scriptsize
    \begin{tabular}{|c|c|c|c|c|c|c|c|}
    \hline
    \multirow{3}{*}{Benchmark Suite} & Offline & \multicolumn{2}{c|}{Single-Stream} & \multicolumn{2}{c|}{Server} & \multicolumn{2}{c|}{Multi-Stream}\\
    \cline{2-8}  & \multirow{2}{*}{(sample/s)} & \multirow{2}{*}{(sample/s)} & 90th percentile  & \multirow{2}{*}{(sample/s)} & 99th percentile  & \multirow{2}{*}{(sample/query)} & 99th percentile\\
     & & & latency (ms) & & latency (ms) & & latency (ms) \\
    \hline
    MLHarness & 133 & 118 & 9.2 & 69 & 44 & 5 & 42  \\ \hline
    MLCommons Inference & 315 & 308 & 3.2 & 121 & 14 & 12 & 44 \\
    \hline
    \end{tabular}
    \label{tab:scenario}
\end{table*}

Table \ref{tab:sys} shows the systems used for experiments. The system naming convention follows the rule as the identifier of the CPU types followed by the acronym of the ML/DL framework, and then the identifier of the GPU type if a GPU is used. There are three system instance categories in total. The first category is an Intel desktop-grade CPU system, including system \SYSA{}, \SYSB{}, \SYSC{}, \SYSD{}, and \SYSJ{}. The second category is also an Intel but different desktop-grade CPU system, including system \SYSE{}, \SYSF{}, and \SYSG. The last category is a server-based system using AMD CPUs, including system \SYSH{} and \SYSI{}. We choose different combinations of frameworks, software systems and hardware systems in order to demonstrate the flexibility and scalability of \SYSNAME{} as a harness for benchmarking.

For both sets of experiments, we report the accuracy, the throughput in the offline scenario, and the throughput and 90 percentile latency in the single-stream scenario. The accuracy is defined differently for each modalities, including the top-1 accuracy for image classification, mAP scores for object detection, F1 scores for question answering, and mean DICE scores for medical image 3D segmentation. As defined in MLCommons Inference \cite{reddi2019mlperf}, the offline scenario represents applications where all data are immediately available and latency is unconstrained, such as photo categorization; on the contrary, the single-stream scenario represents a query stream with sample size of 1, reflecting applications requiring swift responses, such as real time augmented reality. In order to facilitate the comparison between these two scenarios, we fix the batch size of the inferences to 1 in all experiments. This helps to demonstrate how scenarios can affect the throughput of models.

\SYSNAME{} is also capable of reporting results of the other two scenarios as defined by MLCommons Inference \cite{reddi2019mlperf}, which are the server and the multistream scenarios. The server scenario represents applications where query arrival is random and latency is important, such as online translation. The multistream scenario represents application with a stream of queries, where each query consists of multiple inferences, such as multi-camera driver assistance. We demonstrate that \SYSNAME{} is able to report the results of these two scenarios by running \texttt{ResNet50} on 9800-ORT-RTX.

Note that, because of the limited access to data-center scale systems, we are not able to develop and conduct experiments for the rest of the two models as provided by MLCommons Inference, which are \texttt{DLRM} for recommendation system and \texttt{RNNT} for speech recognition. But we believe our methodologies as discussed can be easily extended for the two models.

\begin{table*}[t]
    \centering
    \caption{\SYSNAME{} reported results for models in MLCommons Inference}
    \scriptsize
    \begin{tabular}{|c|c|c|c|c|c|}
    \hline
    \multirow{2}{*}{Model} & \multirow{2}{*}{System} & \multirow{2}{*}{Accuracy} & \multirow{2}{*}{Offline (sample/s)} & \multicolumn{2}{c}{Single-Stream} \vline \\
    \cline{5-6} & & & & (sample/s) & 90th percentile latency (ms) \\
    \hline
    \multirow{6}{*}{MLPerf ResNet50} & \SYSA & Top1: 76.452\% & 133 & 118 & 9.2 \\
    \cline{2-6} & \SYSC & Top1: 76.456\% & 63 & 62 & 16 \\
    \cline{2-6} & \SYSE & Top1: 76.456\% & 118 & 116 & 9.2 \\
    \cline{2-6} & \SYSG & Top1: 76.456\% & 20 & 20 & 57 \\
    \cline{2-6} & \SYSH & Top1: 76.456\% & 159 & 146 & 6.7 \\
    \cline{2-6} & \SYSI & Top1: 76.456\% & 202 & 154 & 6.4 \\
    \hline
    \multirow{6}{*}{MLPerf MobileNet} & \SYSA & Top1: 71.676\% & 196 & 160 & 6.5 \\
    \cline{2-6} & \SYSC & Top1: 71.676\% & 61 & 58 & 23 \\
    \cline{2-6} & \SYSE & Top1: 71.676\% & 188 & 159 & 6.6 \\
    \cline{2-6} & \SYSG & Top1: 71.676\% & 24 & 24 & 44 \\
    \cline{2-6} & \SYSH & Top1: 71.666\% & 358 & 270 & 3.7 \\
    \cline{2-6} & \SYSI & Top1: 71.676\% & 382 & 319 & 3.2 \\
    \hline
    \multirow{6}{*}{MLPerf SSD MobileNet 300x300} & \SYSA & mAP: 23.172\% & 35 & 32 & 37 \\
    \cline{2-6} & \SYSC & mAP: 23.173\% & 28 & 28 & 37 \\
    \cline{2-6} & \SYSE & mAP: 23.173\% & 30 & 28 & 41 \\
    \cline{2-6} & \SYSG & mAP: 23.173\% & 13 & 13 & 78 \\
    \cline{2-6} & \SYSH & mAP: 23.170\% & 20 & 20 & 52 \\
    \cline{2-6} & \SYSI & mAP: 23.173\% & 18 & 18 & 57 \\
    \hline
    \multirow{6}{*}{MLPerf SSD ResNet34 1200x1200} & \SYSA & mAP: 19.961\% & 20 & 19 & 54 \\
    \cline{2-6} & \SYSC & mAP: 19.955\% & 1.4 & 1.7 & 816 \\
    \cline{2-6} & \SYSE & mAP: 19.955\% & 16 & 15 & 66 \\
    \cline{2-6} & \SYSG & mAP: 20.215\% & 1.4 & 1.4 & 704 \\
    \cline{2-6} & \SYSH & mAP: 19.957\% & 23 & 21 & 54 \\
    \cline{2-6} & \SYSI & mAP: 19.955\% & 14 & 13 & 95 \\
    \hline
    \multirow{8}{*}{MLPerf BERT} & \SYSA & F1: 90.874\% & 41 & 38 & 27 \\
    \cline{2-6} & \SYSB & F1: 90.881\% & 21 & 18 & 67 \\
    \cline{2-6} & \SYSC & F1: 90.874\% & 2.2 & 2.5 & 487 \\
    \cline{2-6} & \SYSD & F1: 90.874\% & 0.86 & 0.85 & 1305 \\
    \cline{2-6} & \SYSE & F1: 90.874\% & 30 & 29 & 35 \\
    \cline{2-6} & \SYSF & F1: 90.874\% & 27 & 26 & 39 \\
    \cline{2-6} & \SYSH & F1: 90.879\% & 92 & 78 & 15 \\
    \cline{2-6} & \SYSI & F1: 90.874\% & 29 & 29 & 37 \\
    \hline
    \multirow{2}{*}{MLPerf 3D-UNet} & \SYSH & mean: 0.85300 & 0.043 & 0.045 & 22655 \\
    \cline{2-6} & \SYSI & mean: 0.85300 & 0.045 & 0.045 & 22194 \\
    \hline
    \end{tabular}
    \label{tab:inML}
\end{table*}

\subsection{Results of Models in MLCommons Inference}
In this set of experiments, we demonstrate the capability of \SYSNAME{} on reporting MLCommons Inference \cite{reddi2019mlperf} defined metrics, by benchmarking representative MLCommons Inference models with a variety of systems. 

Table \ref{tab:scenario} shows the various MLCommons Inference defined experimental results of \texttt{ResNet50} produced by \SYSNAME{} on system \SYSA{}. From the results, we observe that using \SYSNAME{}, running \texttt{ResNet50} on such a target system is able to classify 133 images per second, respond to a query of one image in less than 9.2 ms in 90\% of the time if the queries are received contiguously, respond to a query of one image in less than 44 ms in 99\% of the time if the queries are received following the Poisson distribution with an average of 69 queries per second, and respond to a query of five images in less than 42 ms in 99\% of the time if the queries are received contiguously. Note that the number of queries per second in the server scenario and the number of samples per query in the multistream scenarios are tunable parameters for the system to meet the latency requirements. 

\begin{figure}[t]
    \begin{center}
        \includegraphics[scale=0.4]{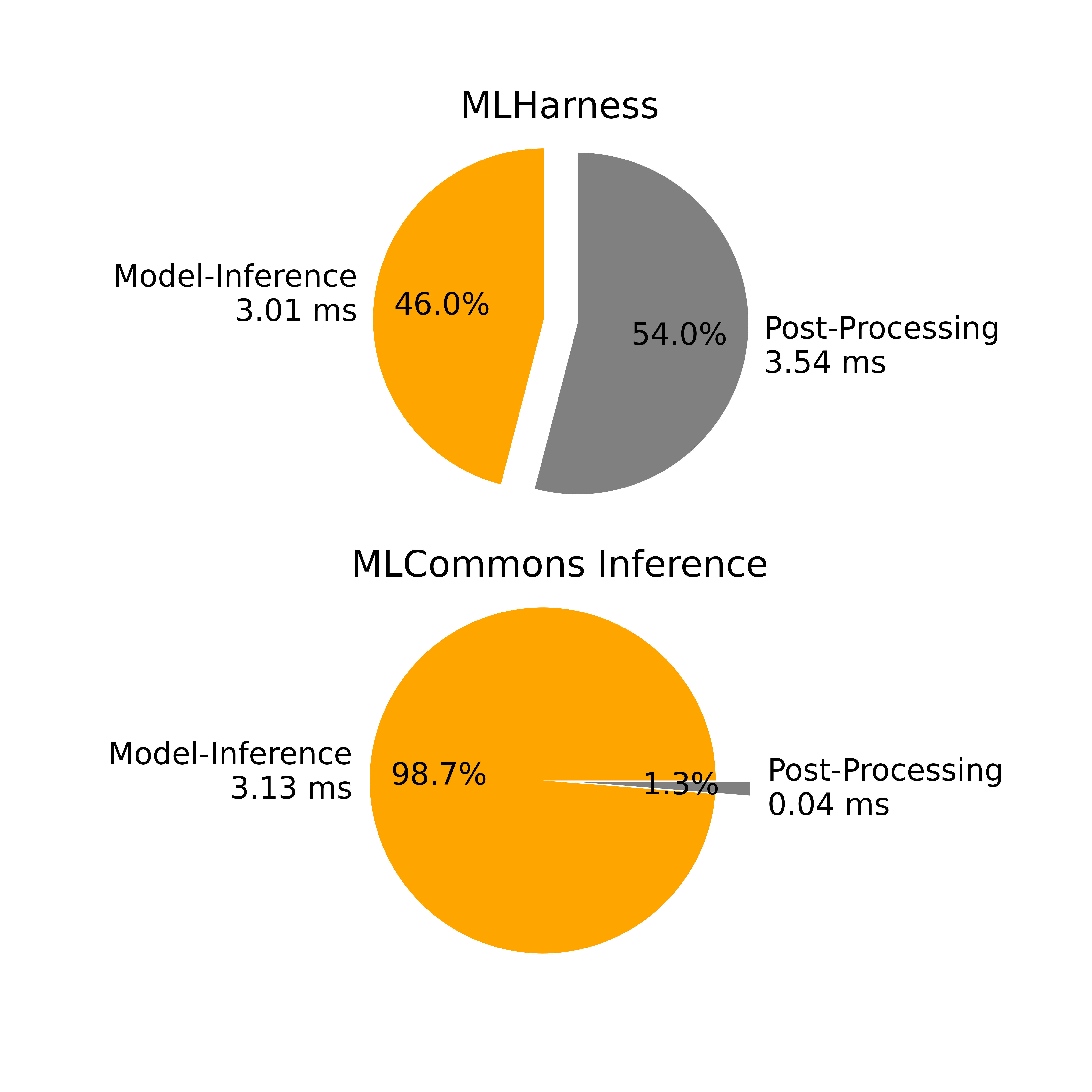}
    \end{center}
    \caption{Break down execution time into model-inference time and post-processing time for \SYSNAME{} and MLCommons Inference running Offline scenario with ResNet50 and a single input on \SYSA{}.}
    \label{fig:pie}
\end{figure}

We also run the same set of experiments on \SYSA{} using the original MLCommons Inference flows, as shown in Table \ref{tab:scenario}. The results show that MLCommons Inference performs two to three times better than \SYSNAME{}. In order to investigate the discrepancy that \SYSNAME{} has a worse performance than MLCommons Inference, we take the Offline scenario as an example and break down the execution time into two parts, including (1) model-inference time for the interval between the model receives pre-processed input tensors and returns output tensors and (2) post-processing time involving generating MLCommons Inference defined format. As Figure \ref{fig:pie} shows, while both \SYSNAME{} and MLCommons Inference spend nearly the same amount of time on model inference, the much higher latency for \SYSNAME{} to post-process data make it hard to achieve the same performance as reported by MLCommons Inference. The underlying reason of this high latency in \SYSNAME{} is due to the aggregated data transferring time between different languages, as data need to be moved several times among ML/DL frameworks, post-processing interfaces and wrappers, while it is not the case for MLCommons Inference since once the inference is done, data always reside in Python. One way to mitigate this high latency is to further optimize \SYSNAME{} for MLCommons Inference by responding directly to the LoadGen in the post-processing function instead of transferring data back to \SYSNAME{} and then reporting to MLCommons Inference suite through wrappers between different languages.  

Table \ref{tab:inML} shows the experimental results of \texttt{ResNet50} and \texttt{MobileNet} for image classification, \texttt{SSD MobileNet 300x300} and \texttt{SSD ResNet34 1200x1200} for object detection, \texttt{BERT} for question answering, and \texttt{3D-UNet} for medical image 3D segmentation. All of these models are provided by MLCommons Inference and can be found at its GitHub page \cite{MLCommonsInferenceGithub}. 

\begin{figure*}[t]
    \begin{center}
        \includegraphics[scale=0.5]{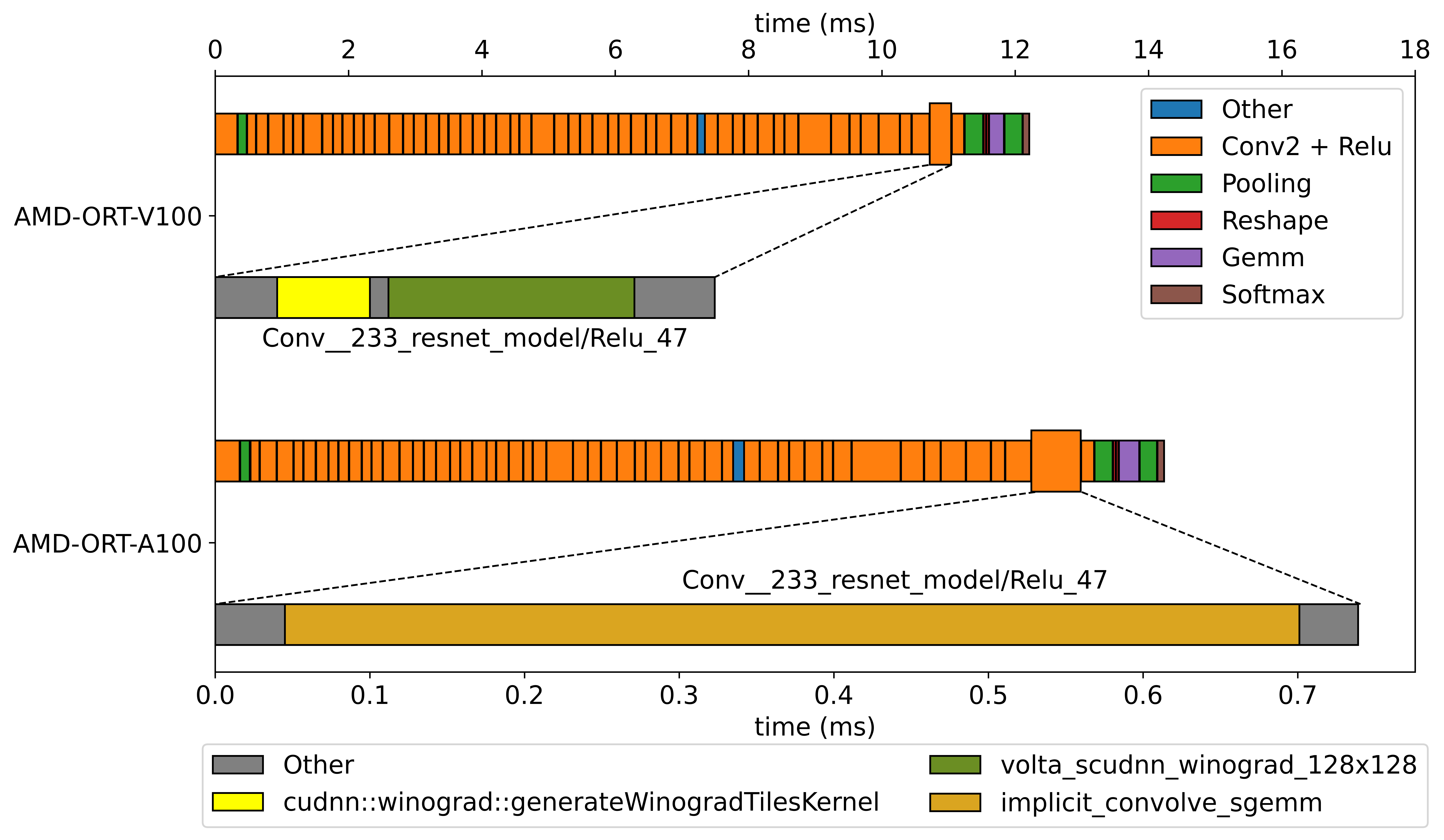}
    \end{center}
    \caption{Performance of ResNet50 with batch size one across systems AMD-ORT-V100 and AMD-ORT-A100 at both layer and kernel (sub-layer) granularity levels, respectively. The axis on the top is the duration to execute each layer in the model, while the axis at the bottom is the duration to execute kernels of the second to the last Conv2 + Relu layer.}
    \label{fig:xspanal}
\end{figure*}

An interesting observation from Table \ref{tab:inML} is that the throughput of \texttt{ResNet50} on system AMD-ORT-V100, which is 202 samples per second, is higher than that on system AMD-ORT-A100, which is 159 samples per second. This seems to be counter-intuitive as the A100 GPU is two generations newer than the V100 GPU, hence the A100 GPU is supposed to have better performance than V100. With MLCommons' inference methodology alone, we are not able to figure out the reason of this ``seemingly abnormal'' behavior. This is the place for \SYSNAME{} to shine with its extended MLModelScope capabilities. Leveraging the across-stack profiling and analysis capabilities from MLModelScope, we are able to align framework-level spans from the ONNX Runtime profiler and the library-level spans from CUDA Profiling Tools Interface, and capture the detailed view of this strange behavior by delving deeper into the results. Figure \ref{fig:xspanal} shows the performance of \texttt{ResNet50} with batch size one across the system AMD-ORT-V100 and system AMD-ORT-A100 at both the layer and the kernel (sub-layer) granularity levels, respectively. At the layer granularity, we observe that the end-to-end inference time on system AMD-ORT-V100 is indeed shorter than that on AMD-ORT-A100, and the reduced runtime mainly comes from the shortened runtime of many Conv2+ReLu layers (in orange color). For example, by focusing only on the second to the last
Conv2+ReLu layer, we see that the duration on system AMD-ORT-A100 is almost twice as large as the duration on system AMD-ORT-V100. By zooming into
that particular layer at the kernel level granularity, we
quickly realize that the two systems have executed different GPU kernels. For system AMD-ORT-V100, there are two major kernels, i.e., cudnn::winograd::generateWinogradTilesKernel and volta\_scudnn\_winograd\_128x128. In contrast, for system AMD-ORT-A100, there is only one major kernel, i.e., implicit\_convolve\_sgemm. We suspect that this discrepancy in performance is mainly due to the less optimized kernel selection algorithm offered by the newer system CUDNN library (v8.1) for A100 GPUs than for V100 GPUs. This further validates
the importance of full-stack optimization for system performance.

In summary, we show that  \SYSNAME{} is capable of reporting MLCommons Inference defined metrics by encapsulating MLModelScope \cite{DBLP:journals/corr/abs-2002-08295} as an easy-to-use black box into MLCommons Inference, and that our harness system is able to benchmark models that are not supported in MLModelScope, including \texttt{BERT} for question answering and \texttt{3D-UNet} for medical image 3D segmentation, with the help of the new interfaces for user-defined pre-processing and post-processing functions. Moreover, as \SYSNAME{} is built on top of MLModelScope, we are able to utilize its across-stack profiling and analysis capabilities to align the information across the ML/DL framework level and the accelerating library level, and pinpoint critical distinctions between models, frameworks, and system.

\begin{table*}[ht]
    \centering
    \caption{\SYSNAME{} reported results for models beyond MLCommons Inference using PyTorch and ONNX Runtime as ML/DL frameworks}
    \scriptsize
    \begin{tabular}{|c|c|c|c|c|c|}
    \hline
    \multirow{2}{*}{Model} & \multirow{2}{*}{System} & \multirow{2}{*}{Accuracy} & \multirow{2}{*}{Offline (sample/s)} & \multicolumn{2}{c}{Single-Stream} \vline \\
    \cline{5-6} & & & & (sample/s) & 90th percentile latency (ms) \\
    \hline
    \multirow{6}{*}{TorchVision AlexNet} & \SYSA & Top1: 56.520\% & 218 & 171 & 6.1 \\
    \cline{2-6} & \SYSB & Top1: 56.516\% & 191 & 154 & 6.8 \\
    \cline{2-6} & \SYSC & Top1: 56.522\% & 86 & 81 & 12 \\
    \cline{2-6} & \SYSD & Top1: 56.522\% & 12 & 12 & 90 \\
    \cline{2-6} & \SYSE & Top1: 56.522\% & 219 & 168 & 6.1 \\
    \cline{2-6} & \SYSF & Top1: 56.522\% & 186 & 152 & 6.9 \\
    \hline
    \multirow{6}{*}{TorchVision ResNet18} & \SYSA & Top1: 69.758\% & 179 & 144 & 7.3 \\
    \cline{2-6} & \SYSB & Top1: 69.756\% & 122 & 113 & 9.6 \\
    \cline{2-6} & \SYSC & Top1: 69.758\% & 128 & 118 & 8.8 \\
    \cline{2-6} & \SYSD & Top1: 69.758\% & 28 & 32 & 42 \\
    \cline{2-6} & \SYSE & Top1: 69.758\% & 175 & 145 & 7.2 \\
    \cline{2-6} & \SYSF & Top1: 69.758\% & 132 & 119 & 9.2 \\
    \hline
    \multirow{6}{*}{TorchVision ResNet34} & \SYSA & Top1: 73.314\% & 142 & 124 & 8.7 \\
    \cline{2-6} & \SYSB & Top1: 73.306\% & 90 & 89 & 12 \\
    \cline{2-6} & \SYSC & Top1: 73.314\% & 72 & 70 & 14 \\
    \cline{2-6} & \SYSD & Top1: 73.314\% & 20 & 19 & 65 \\
    \cline{2-6} & \SYSE & Top1: 73.314\% & 144 & 125 & 8.5 \\
    \cline{2-6} & \SYSF & Top1: 73.314\% & 98 & 93 & 12 \\
    \hline
    \multirow{6}{*}{TorchVision ResNet50} & \SYSA & Top1: 76.130\% & 129 & 115 & 9.4 \\
    \cline{2-6} & \SYSB & Top1: 76.132\% & 76 & 76 & 15 \\
    \cline{2-6} & \SYSC & Top1: 76.130\% & 63 & 61 & 16 \\
    \cline{2-6} & \SYSD & Top1: 76.130\% & 7.9 & 7.7 & 149 \\
    \cline{2-6} & \SYSE & Top1: 76.130\% & 128 & 112  & 9.6 \\
    \cline{2-6} & \SYSF & Top1: 76.130\% & 79 & 78 & 15 \\
    \hline
    \multirow{6}{*}{TorchVision ResNet101} & \SYSA & Top1: 77.374\% & 100 & 89 & 12 \\
    \cline{2-6} & \SYSB & Top1: 77.376\% & 59 & 68 & 22 \\
    \cline{2-6} & \SYSC & Top1: 77.374\% & 36 & 36 & 29 \\
    \cline{2-6} & \SYSD & Top1: 77.374\% & 4.7 & 4.8 & 239 \\
    \cline{2-6} & \SYSE & Top1: 77.374\% & 94 & 87 & 12 \\
    \cline{2-6} & \SYSF & Top1: 77.374\% & 60 & 67 & 22 \\
    \hline
    \multirow{6}{*}{TorchVision ResNet152} & \SYSA & Top1: 78.310\% & 84 & 76 & 15 \\
    \cline{2-6} & \SYSB & Top1: 78.312\% & 46 & 64 & 26 \\
    \cline{2-6} & \SYSC & Top1: 78.312\% & 26 & 26 & 41 \\
    \cline{2-6} & \SYSD & Top1: 78.312\% & 3.5 & 3.5 & 324 \\
    \cline{2-6} & \SYSE & Top1: 78.312\% & 74 & 72 & 15 \\
    \cline{2-6} & \SYSF & Top1: 78.312\% & 52 & 60 & 26 \\
    \hline
    \end{tabular}
    \label{tab:beyondML}
\end{table*}

\begin{table*}[t]
    \centering
    \caption{\SYSNAME{} reported results for models beyond MLCommons Inference using TensorFlow and MXNet as ML/DL frameworks}
    \scriptsize
    \begin{tabular}{|c|c|c|c|c|c|}
    \hline
    \multirow{2}{*}{Model} & \multirow{2}{*}{System} & \multirow{2}{*}{Accuracy} & \multirow{2}{*}{Offline (sample/s)} & \multicolumn{2}{c}{Single-Stream} \vline \\
    \cline{5-6} & & & & (sample/s) & 90th percentile latency (ms) \\
    \hline
    \multirow{2}{*}{VGG16} & \SYSG & Top1: 70.962\% & 9.3 & 9.2 & 115 \\
    \cline{2-6} & \SYSJ & Top1: 72.852\% & 100 & 88 & 11 \\
    \hline
    \multirow{2}{*}{VGG19} & \SYSG & Top1: 71.056\% & 8.7 & 8.6 & 123 \\
    \cline{2-6} & \SYSJ & Top1: 73.814\% & 91 & 82 & 12 \\
    \hline
    \end{tabular}
    \label{tab:beyondMLContd}
\end{table*}

\subsection{Results of Models beyond MLCommons Inference}
Unlike the first set of experiments, which focuses on showcasing the success of \SYSNAME{} in orchestrating MLCommons Inference \cite{reddi2019mlperf} way of benchmarking MLCommons Inference models, the second set of experiments illustrates how to make use of the exchange specification and the across-stack profiling and analysis tool in MLModelScope \cite{DBLP:journals/corr/abs-2002-08295} to facilitate developments and comparisons of ML/DL innovations in the context of MLCommons Inference methodologies.

Table \ref{tab:beyondML} shows a sample of six models to demonstrate how easy it is to use \SYSNAME{} to scale the MLCommons Inference way of benchmarking of various models beyond MLCommons Inference on a variety of system configurations. In this particular example, these
results further show
the relationships among the depth of the convolutional neural networks, the accuracy, and the throughput. The six models are \texttt{AlexNet} along with five models from the \texttt{ResNet} family. All of these models are from TorchVision \cite{TorchVision}, where the implementation details and the reference accuracy can be found at its GitHub page \cite{TorchVisionGithub}. Again, the success of importing these models into \SYSNAME{} using the exchange specification is validated by the accuracy results, where all of them are within at least  99\% of the reference accuracy as stated by TorchVision. In addition, the pre-processing and post-processing functions in the exchange specification can be regarded as a reusable component because these models share the same pre-processing and post-processing steps.

\begin{figure}[t]
    \begin{center}
        \includegraphics[scale=0.25]{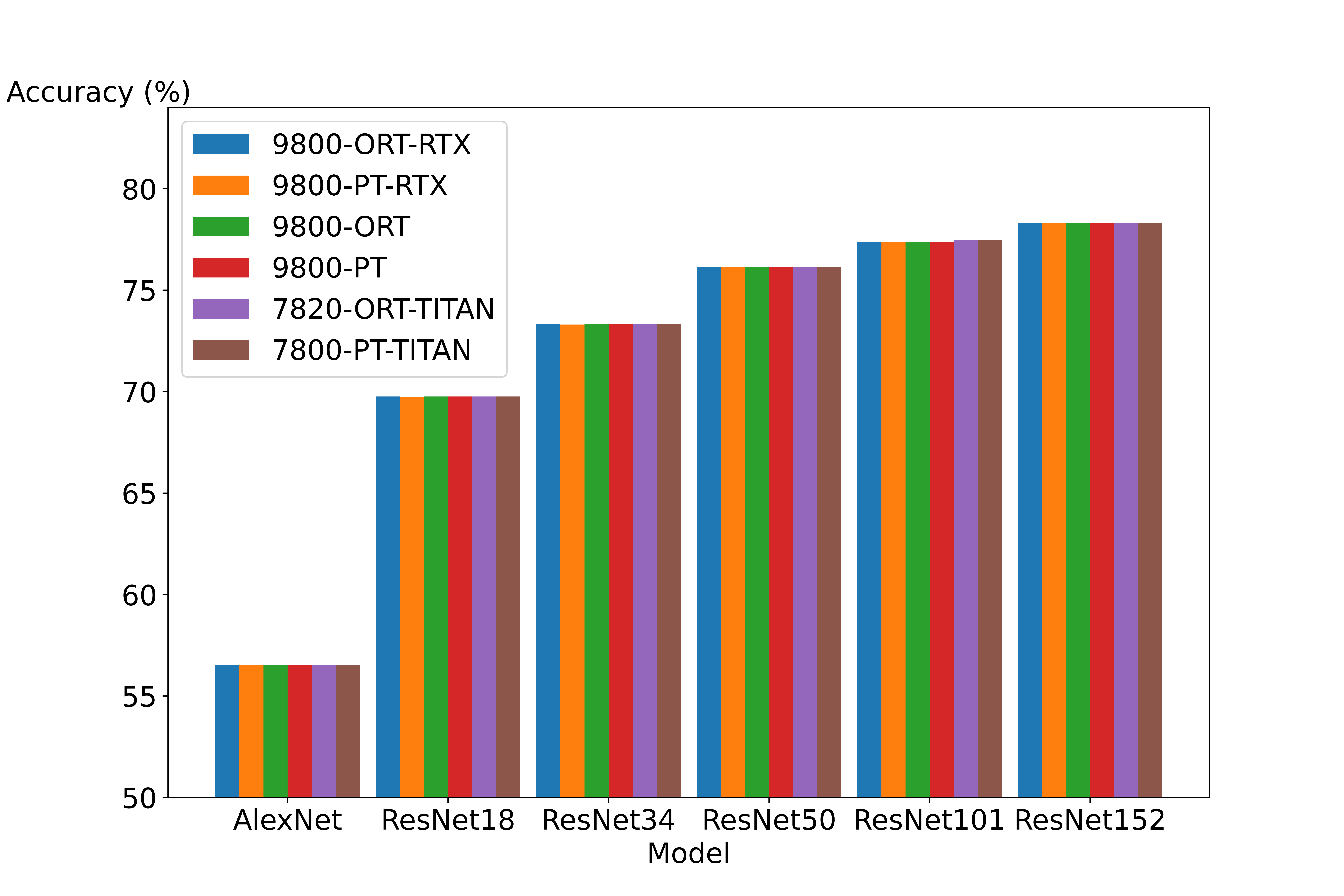}
    \end{center}
    \caption{Accuracy of Models in Different Systems}
    \label{fig:accuracy}
\end{figure}

Figure \ref{fig:accuracy} compares the accuracy of the six convolutional neural networks in systems \SYSA{} to \SYSF{} as listed in Table \ref{tab:sys}. The models are placed in increasing order of depth from left to right, with \texttt{ResNet152} being the deepest. As expected, there is no huge variance on the accuracy across systems, but the deeper the convolutional neural network is, the more accurate the model is.

\begin{figure}[t]
    \begin{center}
        \includegraphics[scale=0.25]{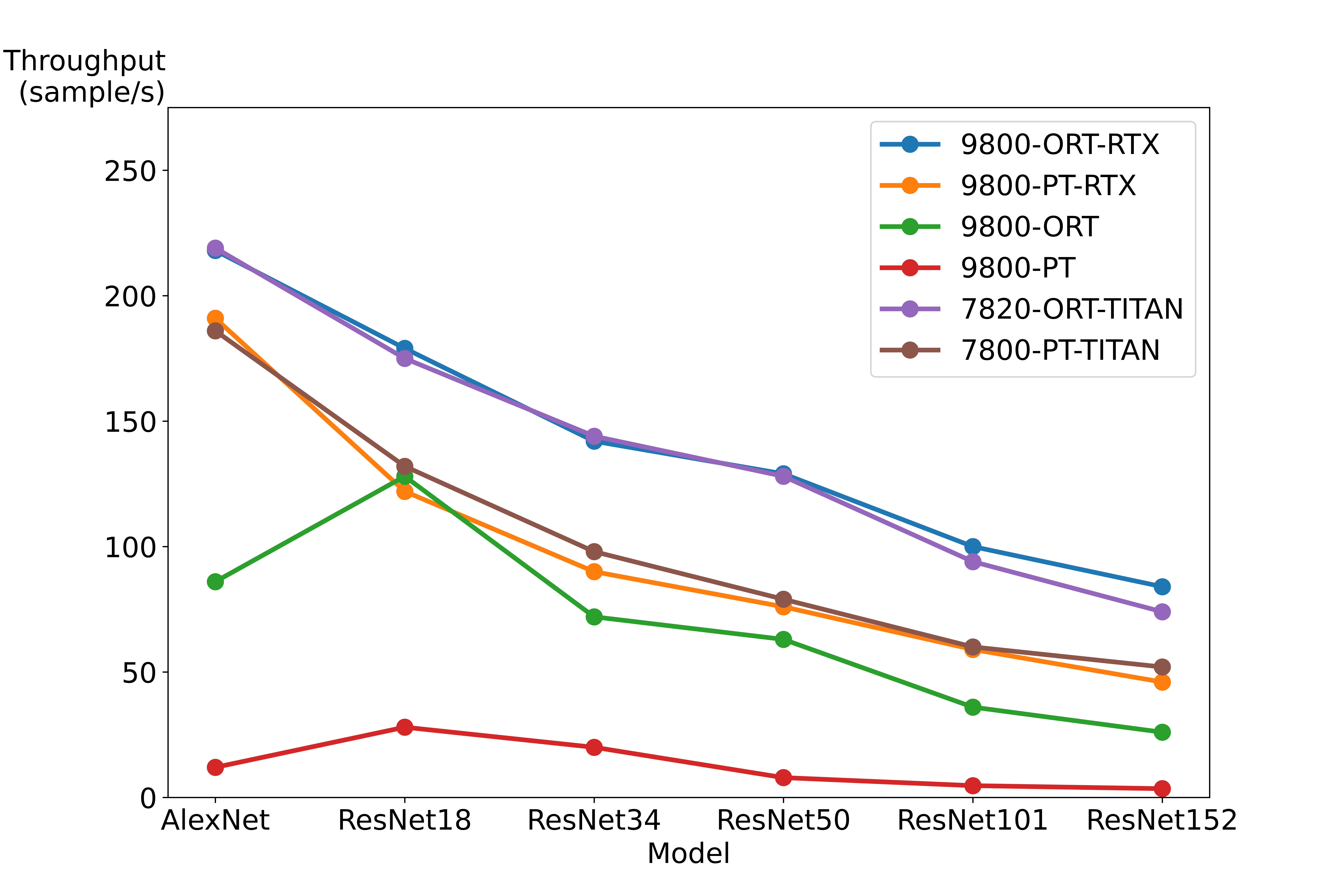}
    \end{center}
    \caption{Offline Throughput of Models in Different Systems}
    \label{fig:throughput}
\end{figure}

Figure \ref{fig:throughput} compares the throughput of the six convolutional neural networks in system \SYSA{} to \SYSF{} as listed in Table \ref{tab:sys}. From Figure \ref{fig:throughput}, we observe that there is a trend that the deeper the convolutional neural network is, the lower the throughput it has. However, for the two systems \SYSC{} and \SYSD{}, which are the two system configurations without GPUs, are not following the trend when comparing \texttt{AlexNet} and \texttt{ResNet18}. As our \SYSNAME{} is built on top of MLModelScope, we then use the across-stack profiling and analysis tool, XSP \cite{DBLP:journals/corr/abs-1908-06869}, to identify the bottleneck. Table \ref{tab:XSP} shows the top three most time-consuming layers of \texttt{AlexNet} and \texttt{ResNet18} identified by XSP on system \SYSD{}, which has no GPU support and has PyTorch as the ML/DL framework. It clearly points out that the bottleneck of \texttt{AlexNet} is from matrix multiplications of fully connected layers. Although there is also a fully connected layer in \texttt{ResNet18} as recorded by XSP, its size is 512 by 1000, which is much smaller than the largest one in \texttt{AlexNet}, whose size is 4096 by 4096.

Although in Table \ref{tab:beyondML}, we use the same implementations of models by converting PyTorch models to ONNX formats that can be used in ONNX Runtime, it is also valuable to compare the same structure of model with different implementations and training processes. Table \ref{tab:beyondMLContd} shows the experiments on the models from the \texttt{VGG} family using TensorFlow and MXNet as ML/DL frameworks, where the models for TensorFlow can be found at TensorFlow Model Graden \cite{tensorflowmodelgarden2020} and the models for MXNet can be found at GluonCV \cite{GluonCV}. From Table \ref{tab:beyondMLContd}, we can observe that the accuracy is different between implementations of the same model, which further illustrates the difficulty of model reproducibility. This also shows how flexible \SYSNAME{} is in terms of running scalable
benchmarking across different
combinations of models and frameworks by utilizing the extended exchange specification as discussed in this work, and
how scalable experimentation
helps to identify common issues convincingly.

In summary, these exemplar experiments as discussed in this section show not only that it is easy to add models into \SYSNAME{} by utilizing the extended exchange specification and to report MLCommons Inference defined metrics for models that are beyond MLCommons Inference, but also that, with the help of MLModelScope, \SYSNAME{} can easily and scalably compare models and extract critical and detailed information, which is impossible when merely using MLCommons Inference. 

\begin{table}[t]
    \centering
    \caption{The top-3 most time-consuming layers\\ of AlexNet and ResNet18 on system \SYSD{}}
    \scriptsize
    \begin{tabular}{|c|c|c|c|}
    \hline
    \multicolumn{2}{|c|}{AlexNet} & \multicolumn{2}{c|}{ResNet18} \\
    \hline
    Layer Name & Latency (ms) & Layer Name & Latency (ms) \\
    \hline
    aten::mm & 47.99 & aten::maxpool2d & 6.31 \\
    \hline
    aten::mm & 17.99 & aten::convolution & 2.46 \\
    \hline
    aten::mm & 4.13 & aten::convolution & 2.24 \\
    \hline
    \end{tabular}
    \label{tab:XSP}
\end{table}

\subsection{Impact of \SYSNAME{}}
The experimental results above demonstrate the success of \SYSNAME{} in benchmarking ML/DL model inferences by providing an extended exchange specification for researchers to easily plug in their ML/DL innovations and collect a set of well-defined metrics.
One of our near future goals is to further extend \SYSNAME{} to support MLCommons training \cite{mattson2019mlperf}.
Nevertheless, the impact of \SYSNAME{} is not only restricted to ML/DL community. 
Benchmarking, reproducibility, portability, and scalability are important aspects in any computing-related research, such as high performance computing and computational biology. The success of \SYSNAME{} is only a starting point, from which we are aiming for extending the same techniques  to other research domains that utilize heterogeneous computational resources, and providing a scalable and flexible harness system to overcome the similar set of challenges. 

\section{Conclusion}
As ML/DL community is flourishing, it becomes increasingly imperative to standardize a common set of measures for people to benchmark and compare ML/DL models quality and performance on a common ground. In this paper, we present \SYSNAME{}, a scalable benchmarking harness system, to remedy and ease the adoption of ML/DL innovations. Our experimental results show superior flexibility and scalability of \SYSNAME{} for benchmarking and porting, by utilizing the extended exchange specification and reporting community acknowledged metrics. We also show that with the help of \SYSNAME{}, we are able to easily pinpoint critical distinctions between ML/DL innovations, by inspecting and aligning profiling information across stacks.  

\section*{Acknowledgements}
This work is supported in part by IBM-ILLINOIS Center for Cognitive Computing Systems Research (C3SR) - a research collaboration as part of the IBM AI Horizons Network.

\printcredits

\bibliographystyle{cas-model2-names}

\bibliography{cas-refs}

\bio{}
\endbio


\end{document}